\documentclass[a4paper, 10pt, conference]{ieeeconf}
\IEEEoverridecommandlockouts                             
\usepackage{color}
\usepackage{graphicx} 
\usepackage{mathtools}
\usepackage{siunitx}
\usepackage[backend=biber, style=ieee,maxbibnames=99]{biblatex}
\AtBeginBibliography{\footnotesize}
\title{\LARGE \bf Collapse of Straight Soft Growing Inflated \\Beam Robots Under Their Own Weight}

\author{Ciera McFarland and Margaret M. Coad
\thanks{The authors are with the Department of Aerospace and Mechanical Engineering, University of Notre Dame, Notre Dame, IN 46556, USA. {\tt\small \{cmcfarl2,mcoad\}@nd.edu}}%
}

\begin{document}

\maketitle
\thispagestyle{empty}
\pagestyle{empty}

\begin{abstract}

Soft, growing inflated beam robots, also known as everting vine robots, have previously been shown to navigate confined spaces with ease. Less is known about their ability to navigate three-dimensional open spaces where they have the potential to collapse under their own weight as they attempt to move through a space. Previous work has studied collapse of inflated beams and vine robots due to purely transverse or purely axial external loads. Here, we extend previous models to predict the length at which straight vine robots will collapse under their own weight at arbitrary launch angle relative to gravity, inflated diameter, and internal pressure. Our model successfully predicts the general trends of collapse behavior of straight vine robots. We find that collapse length increases non-linearly with the robot's launch angle magnitude, linearly with the robot's diameter, and with the square root of the robot's internal pressure. We also demonstrate the use of our model to determine the robot parameters required to grow a vine robot across a gap in the floor. This work forms the foundation of an approach for modeling the collapse of vine robots and inflated beams in arbitrary shapes.    

\end{abstract}

\section{Introduction} \label{introduction}
Robots can help people by conducting useful tasks in spaces that are either too challenging or too dangerous for humans to enter. Ideally, a robot sent into these spaces would be able to navigate a variety of obstacles and scenarios. Everting vine robots \cite{blumenschein_design_2020}---soft, inflated beam robots that ``grow" or lengthen from their tip---can squeeze through tight spaces \cite{hawkes_soft_2017}, grow across rubble \cite{der_maur_roboa_2021},\cite{mishima_development_2006}, and even climb up vertical shafts \cite{coad_vine_2020}. These robots use their already grown body as a ``stem" from which to support the tip as it continues growing. This allows them to navigate easily over and around obstacles. These robots are made of a soft, flexible, and largely non-stretchable membrane filled with air, which is inverted inside itself and ultimately pushed out of the tip by the air pressure. By lengthening from the tip in this way, the robots are able to advance over the ground without sliding their outer wall relative to the environment \cite{hawkes_soft_2017}, which allows them to move through a space without disturbing it.  

\begin{figure}[tb]
\centerline{\includegraphics[width = \columnwidth]{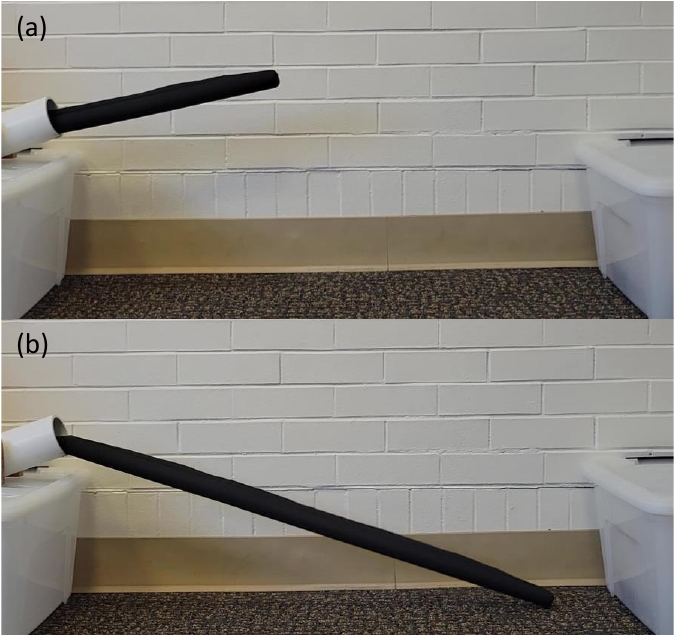}}
\caption{Vine robot collapsing under its own weight. (a) We attempted to grow the vine robot over a gap in the floor, but (b) it collapsed before it arrived. The vine robot's own weight cannot be ignored when considering which tasks are possible.}
\label{gs}
\end{figure}

However, the length to which vine robots can grow unsupported in free space is limited because the hollow body eventually collapses under its own weight. This prevents the robot from achieving certain navigation tasks in open space, such as crossing a gap in the floor, growing across a room from the floor to an open window, or reaching a high space. An example of this failure is shown in Figure~\ref{gs}, where the robot attempts to cross a gap in Figure~\ref{gs}(a) and collapses prior to reaching the other side in Figure~\ref{gs}(b). Vine robots must be able to navigate these types of obstacles in order to be useful in unstructured, open environments.

To enable vine robots to successfully navigate three-dimensional, open-space environments, it is important to understand how they can fail at doing so. Previous work has studied the formation of vine robots into arbitrary shapes \cite{blumenschein_helical_2018},\cite{blumenschein_geometric_2022},\cite{wang_geometric_2022}, as well as the collapse of vine robots due to external loads \cite{luong_eversion_2019},\cite{hwee_evaluation_2021}, environmental contact \cite{greer_obstacle-aided_2018},\cite{greer_robust_2020},\cite{haggerty_characterizing_2019}, and retraction \cite{coad_retraction_2020}. However, no work has studied the collapse of vine robots under their own weight, as their weight has previously been assumed to be negligible. In the inflated beam literature, models have been developed for the collapse of inflated beams under purely axial \cite{fichter_theory_1966} or purely transverse \cite{comer_deflections_1963} external loading, but no work has studied the collapse of inflated beams under off-axis loading or under their own weight. One design used helium to support the weight of a soft robot body in open space \cite{takeichi_development_2017-1}, but this introduces control concerns that may prove challenging over long distances in unstructured spaces.

In this paper, we extend previous inflated beam models to predict when a straight vine robot growing at an arbitrary angle in free space will collapse under its own weight. We validate our model through experiments on physical vine robots at various launch angles, diameters, and internal pressures. We demonstrate the use of our model to select parameters that allow a vine robot to successfully cross a gap. This work is a first step toward determining how an arbitrarily-shaped vine robot will collapse under arbitrary loading. The models developed here are also useful for other inflated beam robots, such as non-everting growing robots \cite{hammond_pneumatic_2017}, heat-welding growing robots \cite{satake_novel_2020}, everting toroidal robots \cite{perez_self-propelled_2022}, and inflated beam truss robots \cite{usevitch_untethered_2020}.

\section{Modeling} \label{modeling}
In this section, we present a model for the length at which a straight vine robot should collapse under its own weight. The literature has identified three main ways in which inflated beams collapse: (1) folding at the base \cite{comer_deflections_1963}, which is typically due to a transverse load (i.e., a load perpendicular to the beam), (2) buckling in the middle \cite{fichter_theory_1966}, which is typically due to an axial compressive load (i.e., a load parallel to the beam), and (3) crushing \cite{le_van_bending_2005}, which is also typically due to an axial compressive load. Crushing only occurs before buckling when the ratio of length to diameter is small, so we will not consider it here. We are interested in how vine robots collapse under their own weight, so the only externally applied force considered here is gravity, which has not been considered in previous work. In the following subsections, we present our models for the collapse length due to folding at the base and buckling in the middle. Similar to the analysis in \cite{coad_retraction_2020}, to create a unified model for any angle of loading, we calculate the collapse length for both cases and assume that whichever happens at a shorter length is the one that will occur.  

\subsection{Folding at the Base}
Folding at the base describes behavior in which an inflated beam folds like a hinge at the last point of support \cite{comer_deflections_1963},\cite{leonard_structural_1960}. The applied moment about the hinge point that causes this folding to occur is referred to as the critical collapse moment. As the critical collapse moment is approached, wrinkles will begin to form on the side of the inflated beam's surface that is opposite where the hinge will occur. When the wrinkles propagate all the way around the beam, the critical collapse moment is reached, and the beam folds. Comer and Levy \cite{comer_deflections_1963} define the critical collapse moment as \begin{equation} \label{eq:1}
M_{collapse} = \frac{P\pi D^3}{8},
\end{equation} where \(P\) is the internal pressure of the inflated beam, and \(D\) is the inflated beam diameter. While this moment was derived for transverse tip loading of inflated beams, it should still apply when the beam is loaded with distributed and/or off-axis loads. Additionally, this collapse moment is determined independently of material properties because the inflated structure is treated as a cantilever beam, which is statically determinate \cite{leonard_structural_1960}. 

\begin{figure}[tb]
\centerline{\includegraphics[width = \columnwidth]{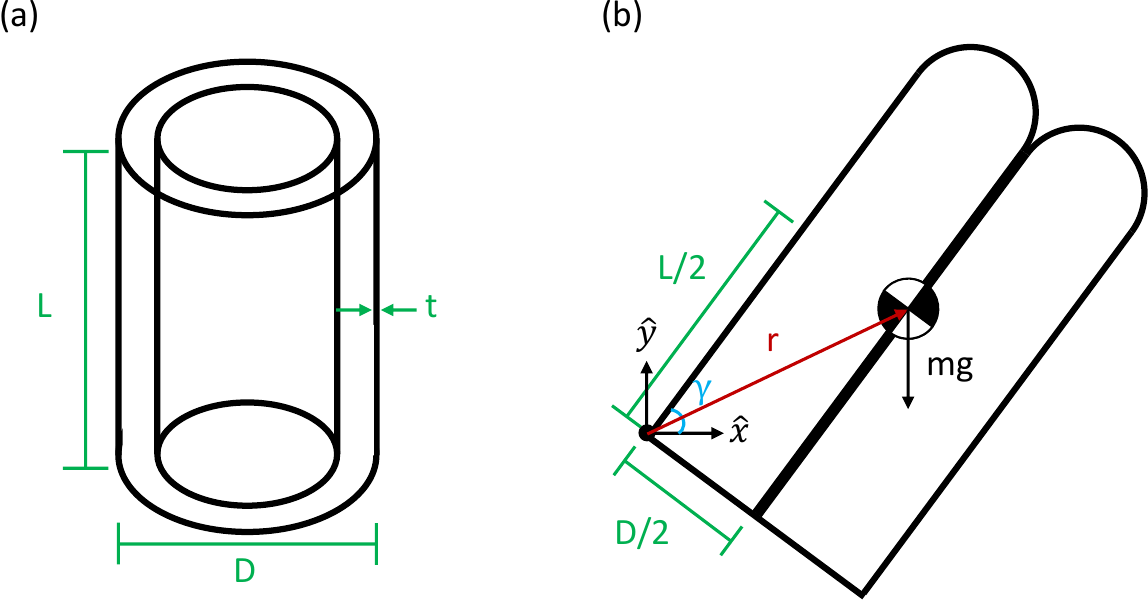}}
\caption{Diagram of parameters used for modeling collapse of a straight vine robot under its own weight. (a) Isometric view of fully grown vine robot. We model the outer wall of the robot body as a circular cylinder of length $L$, diameter $D$, and thickness $t$. (b) Slice through side view of a vine robot. When collapsing due to transverse loading, the robot folds at its base, forming a hinge on one side and wrinkling on the other side. We define the moment about the hinge point created by gravity acting on the center of mass of the robot in terms of the moment arm $r$, diameter $D$, length $L$, robot body mass $m$, gravitational constant $g$, and launch angle $\gamma$ above horizontal. }
\label{diagram}
\end{figure}

To calculate the collapse length of a straight vine robot under its own weight due to folding at the base, we first calculate the moment about the hinge point that is exerted by the robot's own weight. Then, we set that moment equal to the critical collapse moment and solve for the robot length. Figure~\ref{diagram} shows the parameters of our model. As shown in Figure~\ref{diagram}(a), we model the membrane of the fully grown robot as a circular cylinder of length \(L\), diameter \(D\), and thickness \(t\). As shown in Figure~\ref{diagram}(b), we model the pre-grown or growing robot as two layers: the outer cylindrical wall and the inner ``tail" made of the same amount of membrane material, and assumed to be located along the robot's central axis. Gravity acts downward on the center of mass of the robot body (assumed to be in the middle of its length and width) with a magnitude of \(mg\), where \(m\) is the robot body mass and \(g\) is the gravitational constant. When the robot is launched at an angle \(\gamma\) above the horizontal, the force of gravity exerts a moment about the hinge point, which is located on the top side of the base of the robot body. The y-axis points upward, and the x-axis points horizontally, in the plane of the robot body. The vector from the hinge point to the center of mass is \(r\). As the force of gravity acts along the y-axis, the magnitude of the moment due to the robot's weight is

\begin{equation} \label{Mweight}
M_{weight} = mgr_x,
\end{equation}
where \(r_x\) is the x-component of \(r\).

The moment arm \(r_x\) can be calculated by summing the x-components of two vectors: one perpendicular and one parallel to the beam's axis. It is described by \begin{equation} \label{eq:2}
r_x = \frac{D}{2}\sin{\gamma}+\frac{L}{2}\cos{\gamma}.
\end{equation} Here, we define \(m\) as the product of the robot's membrane volume and density. That relationship is \begin{equation} \label{eq:4}
m = 2\pi DtL\rho,
\end{equation} where \(\rho\) is the density of the robot body material. The volume of the outer cylindrical wall is doubled to account for the volume of the inner tail. 

Substituting Equations \ref{eq:2} and \ref{eq:4} into Equation \ref{Mweight}, and solving for L using the positive solution of the quadratic formula, we have an equation for the expected collapse length for a straight vine robot due to folding at the base under its own weight, which is described by \begin{equation} \label{eq:5}
L = \frac{-D}{2}\tan{\gamma} + \frac{D\sqrt{t^2\rho^2g^2\sin^2{\gamma}+\frac{1}{2}t\rho gP\cos{\gamma}}}{2t\rho g\cos{\gamma}}.
\end{equation} 

When \(\gamma\) is equal to zero, this simplifies to \begin{equation} \label{eq:6}
L = D\sqrt{\frac{P}{8t\rho g}}.
\end{equation} 

\subsection{Buckling in the Middle}
Buckling in the middle occurs when an inflated beam collapses due to axial compressive loading. The location of buckling is somewhere between the last point of support and the point where the force is applied \cite{fichter_theory_1966}.

Fichter \cite{fichter_theory_1966} models the force to buckle an inflated beam in the middle due to an axial compressive load at its tip as \begin{equation} \label{eq:8}
F_{buckle} = \frac{E\pi^3R^4tP+EG\pi^3R^3t^2}{E\pi^2R^2t+RL^2P+GtL^2},
\end{equation} where \(R\) is the inflated beam radius, and \(E\) and \(G\) are respectively the Young and shear modulus of the beam wall material. However, as this equation is intended for tip-loaded weight, we use $\frac{L}{2}$ in place of \(L\) as we are modeling the weight to be concentrated at the center of mass, halfway up the beam. When the robot is vertical (\(\gamma\) = 90$^{\circ}$), the force of gravity \(mg\) acts downward on the center of mass and is described by \begin{equation} \label{eq:9}
F_{weight} = 2\pi DtL\rho g.
\end{equation} To check for buckling in the middle at a given length, we simply check whether \(F_{weight}\) exceeds \(F_{buckle}\). 

\section{Experiments and Results} \label{experiments}
In this section, we present the setup and results of three experiments that attempt to validate Equations \ref{eq:5}, \ref{eq:8}, and \ref{eq:9} for vine robots collapsing under their own weight. These experiments consider how the collapse length changes with three other quantities in the equations: launch angle, robot body diameter, and pressure.

\begin{figure}[tb]
\centerline{\includegraphics[width = \columnwidth]{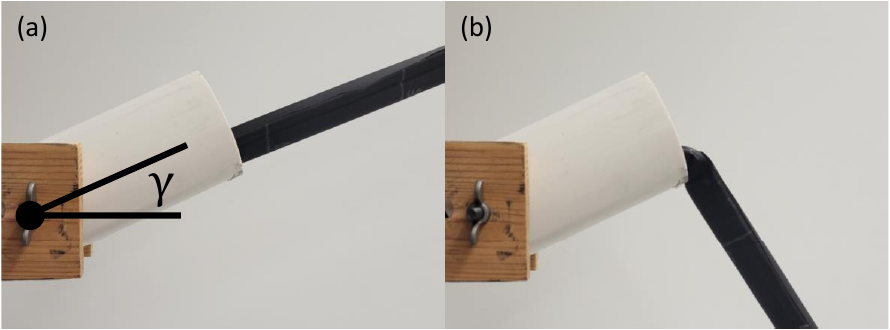}}
\caption{Experimental setup for launching vine robots at various angles. (a) The robot can either be grown or pushed out of the pipe, which can be adjusted to a desired launch angle \(\gamma\) above horizontal. (b) When the robot collapses due to transverse loading, it folds at the edge of the pipe.}
\label{collapse}
\end{figure}

For all experiments, the robot was launched at a consistent and adjustable angle using the launcher setup shown in Figure~\ref{collapse}, which consists of a pipe with an inner diameter of 6.27~cm attached to two wooden boards with threaded screws tightened by wingnuts. This setup forces the robot to grow at a specific launch angle ($\gamma$) with a clear last point of support, the edge of the pipe, which is where we expect collapse to occur.

The robot bodies were made from 40 denier TPU-coated heat-sealable ripstop-nylon (extremtextil, Dresden, Germany). We measured the thickness to be 0.031~mm using calipers and found the density to be 2199~kg/m\(^3\) by weighing a measured sample of the material. We experimentally determined the Young's and shear moduli of the material to be 1.3~GPa and 778~MPa respectively through a tensile test. We formed the fabric into a tube that we sealed along its length with an impulse sealer (Jorestech, Sunrise, FL). All robots were also sealed together at the base such that they could not grow, in order to ensure that there were two layers of material along the entire length at all times. The robots were manufactured to have a flap of 1.27~cm of extra fabric on either side of the seam to ensure the heat-sealed tube was the desired diameter. This flap added extra weight, which is factored into our modeled results as additional robot body area concentrated around the central axis. This flap was only incorporated into Equation~\ref{eq:5} and Equation~\ref{eq:9} because it adds weight to the robot, but is not part of the inflated tube, and therefore should not contribute to the tube dimensions used in Equation~\ref{eq:1} and Equation~\ref{eq:8}. We lengthened the robots by slowly pushing them from behind. This tested the model under ideal conditions, i.e., without the dynamic motion caused by growth, and this made it easier and faster to conduct trials, as we did not need to invert the robot each time. We controlled the pressure using a closed-loop pressure regulator (QB3, Proportion-Air, McCordsville, IN). 

\subsection{Collapse Length for Various Launch Angles}

In the first experiment, we measured the length at which a vine robot collapsed at various launch angles. For these trials, we held the diameter constant at 2.43~cm and the pressure constant at 3.45~kPa. We varied the launch angle from \mbox{-65$^{\circ}$}, i.e., steeply downward, to \mbox{90$^{\circ}$}, i.e., straight upward. At angles below \mbox{-65$^{\circ}$}, clear collapse behavior did not occur within our testing space. For example, in a trial done for a launch angle of \mbox{-75$^{\circ}$}, which should cause the robot to fold at the base at 1.10~m, the robot achieved a length of 1.15 m before hitting the floor, and it had not started wrinkle formation. Given that our model predicts an infinite collapse length for a launch angle of -90$^{\circ}$, we take this behavior as the point where the robot is now able to grow to long lengths without collapsing.  

\begin{figure}[tb]
\centerline{\includegraphics[width = \columnwidth]{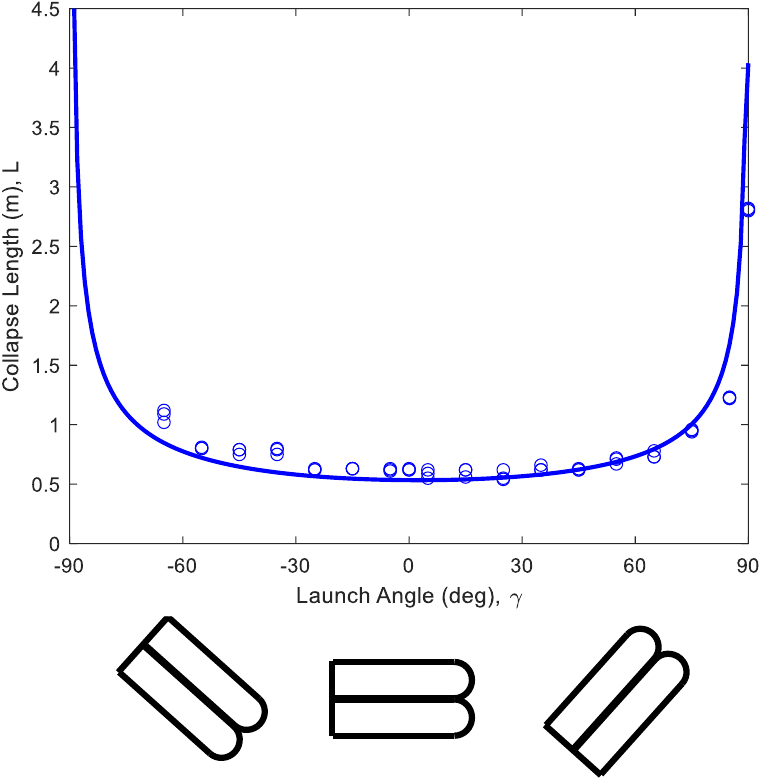}}
\caption{Lengths at which a straight vine robot collapsed at various launch angles. The robot had a diameter of 2.43 cm and an internal pressure of 3.45~kPa. It was launched three times at each angle. Circles are our experimental data, and the solid line is our model. At angles where fewer than three circles appear, the data is overlapping. The experimental collapse lengths follow the same trend as the model, but we see premature collapse at angles greater than \mbox{65$^{\circ}$} and late collapse at angles below the horizontal.}
\label{angleplot}
\end{figure}

For all trials except when the launch angle was \mbox{90$^{\circ}$}, we launched the robot using the setup shown in Figure~\ref{collapse}, elevated such that the robot could not touch the ground without collapsing. Contacting the ground would provide the robot with a new point of support and void the experiment. When the launch angle was \mbox{90$^{\circ}$}, we used our hands to launch the vine robot straight up from the ground in order to help steady it. We also used our hands to create a loose ring around the robot tip. We did this to keep the robot growing at the appropriate angle, as it was prone to wobbling. We made sure to not contact it enough to provide any vertical support. We ran three trials at each angle and deflated and rotated the robot between trials to help prevent material fatigue.

Figure~\ref{angleplot} shows the modeled collapse length for this specific robot across the given angle range (solid line), as well as data for angles ranging from \mbox{-65$^{\circ}$} to \mbox{90$^{\circ}$} (circles). In general, the data follows a similar trend to the model: the collapse length tends to increase with angles of increasing magnitude. However, at positive angles greater than $65^{\circ}$, the robot collapsed at a shorter length than the model predicted, and, at negative angles, the robot collapsed later than anticipated. The model is therefore a good fit for the data, though it is more accurate at positive angles and angles close to the horizontal. 

The robot also collapsed at a shorter length than expected during the purely axial loading test ($\gamma = 90^{\circ}$), which was the only trial where we observed buckling in the middle. While Equation \ref{eq:8} predicted a significantly greater length than was achieved, this assumed a perfect axial load on the robot. Our result is a better fit for models used to determine the critical height of cylinders under their own weight \cite{kanahama_summation_2022}, which can be applied to plants in nature. This suggests we have reached the realistic limitation on collapse length for these robot parameters, even in our controlled setting. Additionally, the model suggests folding at the base should be the dominant behavior for shallower angles. However, this model does not account for the stiffness of the material. The late collapse at these shallower angles may be due in part to the material stiffness supporting the weight past when it is modeled to do so.  

\subsection{Collapse Length for Various Robot Diameters}

\begin{figure}[tb]
\centerline{\includegraphics[width = \columnwidth]{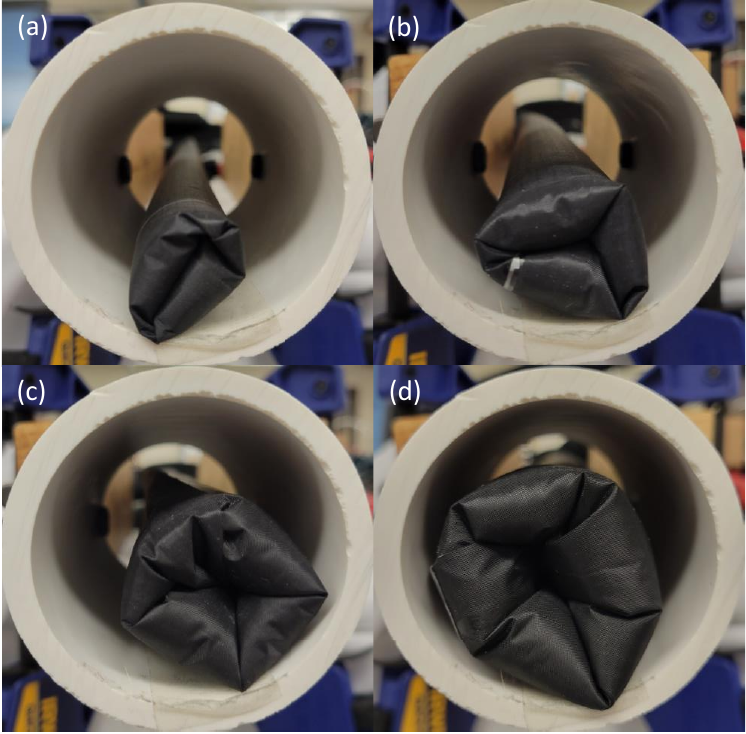}}
\caption{Head-on view of launcher pipe showing the four robot diameters tested. We used robots with diameters of (a) 2.43~cm, (b) 3.24~cm, (c) 4.04~cm, and (d) 4.85~cm.}
\label{diameterphoto}
\end{figure}

In the second experiment, we measured the length at which vine robots of various diameters collapsed. For these trials, we held the launch angle constant at \mbox{0$^{\circ}$} and the pressure constant at 2.07~kPa. The four robots had diameters of 2.43~cm, 3.24~cm, 4.04~cm, and 4.85~cm, as shown in Figure~\ref{diameterphoto}. The four robots are shown in the launcher for clarity, but, as the chosen angle was \mbox{0$^{\circ}$}, the launcher was not needed for this experiment, and we lengthened the robots directly off the table.  As in the previous experiment, we ran three trials at each diameter and deflated and rotated the robot between trials to help prevent material fatigue. This diameter range was chosen to align with the diameters of most existing vine robots in the literature. Smaller diameters would be challenging to fabricate, while larger diameters would be significantly larger than those typically used in the literature.  

\begin{figure}[tb]
\centerline{\includegraphics[width = \columnwidth]{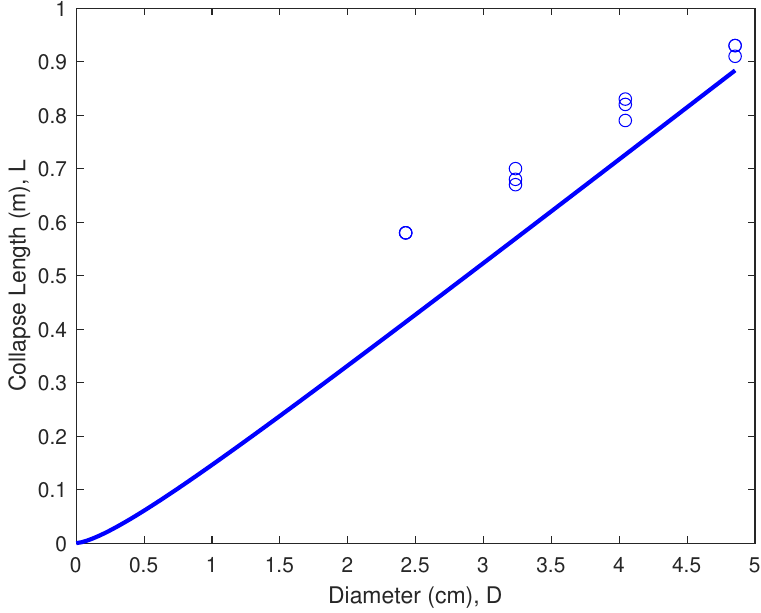}}
\caption{Lengths at which straight vine robots collapsed at various diameters. Circles are our experimental data, and the solid line is our model. We launched the robots horizontally ($ \gamma = 0^{\circ}$) and at an internal pressure of 2.07~kPa three times at each diameter. At diameters where fewer than three circles appear, the data is overlapping. Both the data and the model exhibit a linear relationship, but the slope of the model is slightly steeper than the slope of the data.}
\label{diameterplot}
\end{figure}

Figure~\ref{diameterplot} shows the modeled lengths at which robots should collapse for our chosen diameters (solid line), as well as the data for when they actually collapsed (circles). As predicted by the model, there is a positive linear relationship between collapse length and robot body diameter. The slight non-linearity in the model is because the weight of the extra flap has less of an effect on larger robots. The performance of the smallest vine robot is similar to that in Figure~\ref{angleplot}. The data does have a slightly different slope than the model, but, with increasing diameter, we see better agreement with the model within the range we evaluated. While it is unclear how this trend would continue beyond our range, the model holds well for diameters common in the literature. 

\subsection{Pressure to Collapse and Uncollapse Robots of Various Lengths}

\begin{figure}[b]
\centerline{\includegraphics[width = \columnwidth]{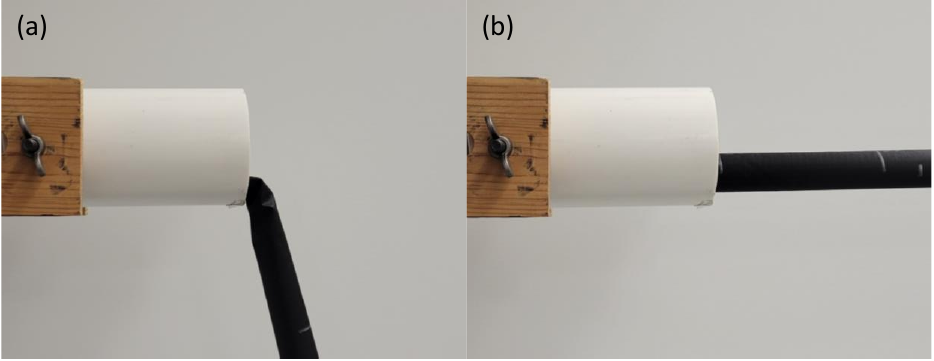}}
\caption{Uncollapsing process. (a) When the robot is fully deflated or at low pressure, it is collapsed. (b) When the robot is inflated to a high enough pressure, it uncollapses and becomes straight.}
\label{uncollapse}
\end{figure}

In the third experiment, we measured the pressure below which a vine robot collapsed and above which it uncollapsed at various lengths. For these trials, we held the diameter constant at 2.43~cm and the launch angle constant at \mbox{0$^{\circ}$}. While we have already shown the collapsing process in Figure~\ref{collapse}, uncollapsing describes what happens when a fully deflated and collapsed robot is inflated such that it becomes completely straight. Figure~\ref{uncollapse} shows the before and after images of this process.

To find the pressure at which the robot collapsed, we launched the robot horizontally out of the pipe while already inflated past its modeled pressure needed to hold the length. We then lowered the pressure in increments of 0.69~kPa until it collapsed, leaving approximately ten seconds between increments to allow the pressure to adjust. Here, the robot was considered collapsed when it folded to an angle of \mbox{-45$^{\circ}$}. This boundary was implemented as the robot often did not fold all the way to \mbox{-90$^{\circ}$} for these trials, due in part to the way the robot folds over the pipe when it is not moving. In Figure~\ref{uncollapse}, the fold is slightly inside the pipe, causing the robot body to angle slightly outward from the pipe even without any pressure. To uncollapse the robot, we launched the desired length completely deflated and collapsed, and then increased the pressure in 0.69~kPa increments, again allowing ten seconds for the pressure to adjust. We ran three trials of each scenario and deflated and rotated the robot between trials. 

\begin{figure}[tb]
\centerline{\includegraphics[width = \columnwidth]{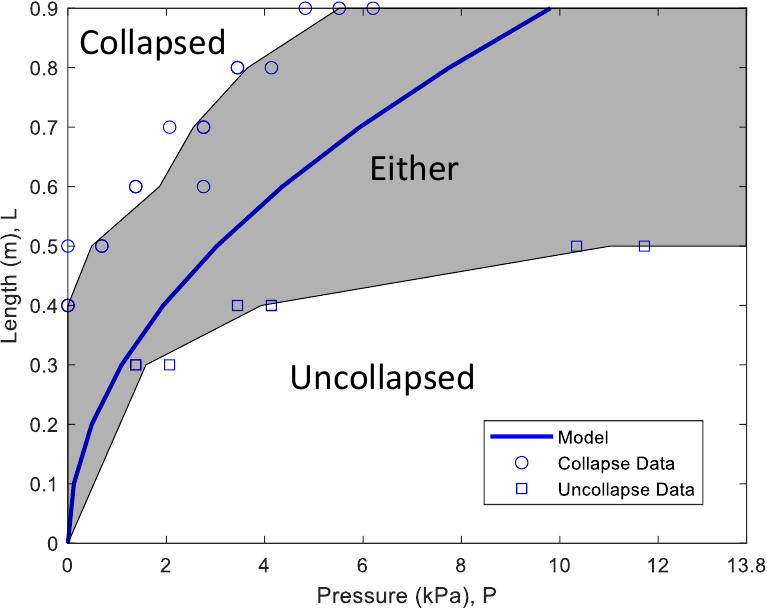}}
\caption{Internal pressure at which a straight vine robot collapsed and uncollapsed at various lengths. We launched the robot horizontally, and it had a diameter of 2.43~cm. Circles are the experimental data for the pressure below which the robot collapsed, squares are the experimental data above which the robot uncollapsed, and the solid line is our model. We ran every collapse and uncollapse scenario three times. In trials where fewer than three data points appear, the data is overlapping, except the noted trial of uncollapsing at 0.5 m, where the robot did not uncollapse within our pressure boundaries. The shaded region, bounded by the average pressure result of each scenario, denotes an area of uncertainty where the robot may be collapsed or uncollapsed, depending on its state upon entering the region. Pressures lower than the predicted one can still support the robot if it starts uncollapsed. However, pressures higher than the predicted one are needed to uncollapse the robot if it starts collapsed.}
\label{pressureplot}
\end{figure}

We varied the length between 0.3~m and 0.9~m to capture many data points in a range we have found in preliminary trials to behave predictably. Lengths of 0.1~m and 0.2~m were excluded because we have not found the model to be accurate for significantly short lengths or low pressures. Even at 0.3~m, it was not possible to gather collapse data, as the stiffness of the material was able to support that length even when the pressure in the robot was 0 kPa. We also did not permit trials where the pressure to support the robot exceeded 13.8~kPa, as this risked bursting the robot. The result of this is there is no uncollapse data for lengths exceeding 0.5~m. One trial for 0.5~m did not uncollapse in our pressure range, resulting in there only being two data points at that length. 

Figure~\ref{pressureplot} shows the modeled length at which the robot should collapse for a certain pressure (solid line). It also shows the data for collapse trials (circles) and uncollapse trials (squares). The collapse data follows the same general trend as the model. However, a robot of a given length can be supported by less pressure than modeled. Additionally, more pressure than modeled is needed to uncollapse the robot. The existence of different collapse and uncollapse pressures means that, in the shaded region, whether the robot is collapsed or uncollapsed depends on its state upon entering the shaded region. This region starts out narrow at short lengths and low pressures, but widens as the length and pressure of the robot increases. Ideally, we would seek to narrow this region to ensure we can use longer length robots without encountering our pressure restrictions. Once again, we believe this discrepancy is caused by the stiffness of the material, which is not accounted for in the folding at the base model. The support of the material is especially clear at shorter lengths, where the robot was able to support itself even without any pressure. It is likely this makes it easier for the robot to stay in its starting position, be that collapsed or uncollapsed.

\section{Discussion} \label{discussion}
In this section, we discuss additional observations made during data collection that do not directly appear in the results, as well as the implications of our results for selecting robot parameters to achieve a desired task.

\subsection{Deflection}
This paper specifically considers when and how collapse occurs. However, when looking at the collapse length for various launch angles, it was clear that the robot only remained straight while it was sufficiently short. As the robot lengthened, it began to deflect downward. If the robot needs to grow more than a short distance, it must be launched at a steeper angle than if it grew completely straight in order to account for this deflection.

We chose to largely ignore this effect for our study because we were only interested in the point of critical failure. However, it is possible that deflection resulted in some of the unexpected collapse data, given that it forces the center of mass to be at a different location than modeled. This would make the expected collapse length shorter at angles greater than \mbox{0$^{\circ}$}, and longer at angles less than \mbox{0$^{\circ}$}, which aligns with some of the unexpected behavior we observed. In previous work, deflection has been measured directly \cite{hwee_evaluation_2021} or used to determine a practical failure point \cite{luong_eversion_2019} using the same foundational equations as our model. In order for this model to be applied for detailed path-planning of the robot tip location, deflection would need to be accounted for, but our current model is close enough for some tasks.     

\subsection{Flap Effects}
We fabricated these vine robots in such a way that they have an extra flap of material along the seam that ran the length of the body. We did this to aid in the heat-sealing process and ensure the body of the robot was the desired diameter. However, the flap added additional weight along the length of the robot. We factored this weight into the model by adding the weight of the flap as if it was on the robot's center axis to the weight of the outer wall and tail. However, we did not account for the flap's location along the circumference. For all trials, we deflated and rotated the robot between tests to help prevent the material from repeatedly collapsing in the same spot. This includes collapsing prematurely at a location that the robot previously collapsed at. We were able to easily observe this rotation by noting the circumferential location of the flap. The location of the flap was especially impactful when exploring the pressure to collapse and uncollapse a robot of a given length. 

We observed that the robot was more likely to collapse at a shorter length than predicted by the model if the flap was along the side of the robot, but this was not guaranteed. The robot was often able to better withstand collapse if the flap was on the bottom. As a result, we conducted most of the pressure trials with the flap at or near the top of the robot body, since this produced the most consistent and repeatable behavior. The effect of circumferential location did not seem to be as prevalent in the other trials. This odd behavior at certain circumferential locations suggests that the heat seal or the extra fabric is preventing wrinkles from occurring as they would on a beam that has been fabricated into a tube without a seal along its edge. The effect is likely lessened when the flap is at the top because in that case, collapse is already occurring when the wrinkles reach the heat seal. Modeling this effect could allow more precise determination of collapse length for all initial robot orientations, but our current model is close enough for some tasks.

\begin{figure*}[tb]
\centerline{\includegraphics[width = \textwidth]{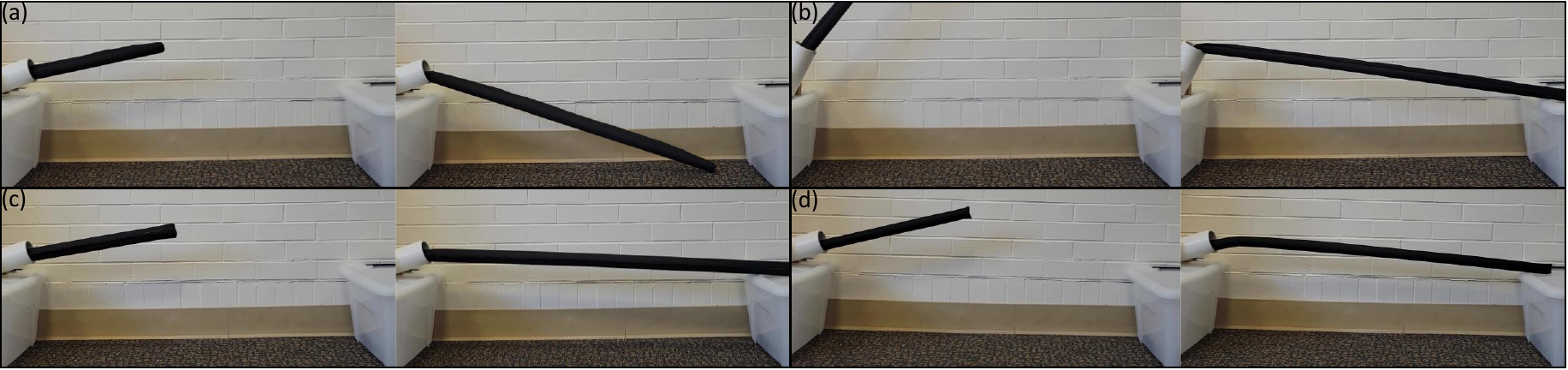}}
\caption{Demonstration of using our model to choose the robot parameters to cross a gap before the robot collapses under its own weight. (a) Our basic robot (launch angle \mbox{20$^{\circ}$}, diameter 3.24~cm, internal pressure 4.14~kPa) begins growing across a gap of 0.95~m, but it collapses before it reaches the other side. As predicted by our model, either (b) increasing the launch angle to $65^{\circ}$, (c) increasing the diameter to 4.04~cm, or (d) increasing the pressure to 10.34~kPa while holding the other parameters fixed, allows the robot to successfully cross the gap.}
\label{demofig}
\end{figure*}

\subsection{Design Implications}
Our results highlight several important design considerations when approaching a vine robot task. When determining a launch angle, the collapse length does not increase significantly unless very steep angles are used. Therefore, barring height restrictions from the environment, it is better to attempt tasks with very steep angles to have a higher chance of success. When attempting a task involving  negative angles, this becomes more complicated, as the steepest angles deflect but do not exhibit consistent collapse behavior. When attempting to cross a gap and reach a platform at the same or a lower level than the vine robot's starting platform, it may be more reliable to use a steep positive angle to obtain the desired length and then let it collapse onto the platform.

When considering the ideal diameter for a vine robot, our model supports that larger diameters will collapse at longer lengths. However, larger diameter robots have greater restrictions on their operating pressure than robots with smaller diameters, because the hoop stress acting on the robot, which determines how close it is to bursting, also scales linearly with diameter at constant pressure. Robots with larger diameters will also have more difficulty growing through small apertures due to the additional material used to form that larger cross-section.

Increasing the robot's pressure does increase its collapse length, but to a lesser extent than increasing the diameter, and with diminishing returns. Given the restrictions on pressure to avoid bursting the robot, this may not always be a feasible way to increase collapse length. Additionally, the robot can turn more easily when its body is at a lower pressure. Operating at the highest allowable pressure will allow for a greater collapse length, but it may also limit navigational abilities. Finally, the high pressure needed to uncollapse the robot suggests that once the robot has collapsed, it will often be more practical to shorten its length and then regrow it than try to uncollapse it at its original collapse length.

\section{Demonstration} \label{demonstration}

We conducted a demonstration of the use of our model to select the robot parameters to achieve a desired task. The task we chose was for the robot to grow from one platform over a gap to another platform the same height as the first, and our goal was to use our model to choose the launch angle, diameter, and pressure to achieve the task. This is a general version of a scenario the robot might often face in complex environments, where it supports itself in three-dimensional space as it moves from one area of interest to the next. 

For the demonstration, the robot's target was 0.95~m away. To best match a real scenario, the robot grew out of the pipe instead of being manually pushed. To match our model, we fabricated it to be sufficiently long such that the portion of the robot that emerged from the launcher always consisted of two layers of fabric along its length. 

We chose a setup where our initial robot could not succeed, which is shown in Figure~\ref{gs} and Figure~\ref{demofig}(a). This robot had a launch angle of \mbox{20$^{\circ}$}, a diameter of 3.24~cm, and a pressure of 4.14~kPa, and it was modeled to collapse at 0.82~m. Experimentally, it did not succeed in crossing the gap. However, there are several ways that this vine robot or a modified one could successfully complete the task. Our chosen robot should be able to succeed with a different launch angle, diameter, or pressure. Based on our model, a robot is less likely to collapse at a steeper launch angle, larger diameter, and higher pressure, so we increased each parameter independent of the other two and reattempted the task. 

We started by increasing the launch angle to \mbox{65$^{\circ}$}, which our model predicts will produce a robot that will collapse at a length of 1.20~m and fall onto our target. In Figure~\ref{demofig}(b), the robot has grown steeply to some length, and it is ultimately able to collapse at a long enough length to reach the target. We previously tried \mbox{55$^{\circ}$}, which was modeled to collapse at 1.04~m, but the robot unexpectedly collapsed prematurely. Therefore, we chose a very steep angle where the robot was expected to succeed even if it collapsed at a shorter length than modeled as we saw in our experimental trials. 

Next, we returned to our initial angle and instead used a different robot with a diameter of 4.04~cm, which was modeled to collapse at 1.05~m. As predicted, this robot steadily grows at the desired angle before ultimately collapsing onto the target, which can be seen in Figure~\ref{demofig}(c). 

Finally, we returned to our initial robot diameter and launch angle and instead increased its pressure to 10.34~kPa, which was modeled to collapse at 1.31~m. This robot was also able to grow and collapse onto the desired target, which is shown in Figure~\ref{demofig}(d). This significant increase in pressure was made necessary by consistent premature collapse at lower pressures, which may have been due to the flap's circumferential location or the dynamic nature of growth. 

This demonstration shows how we can use our model to help predict successful and unsuccessful robot parameters to achieve a desired task. If we have the opportunity to choose the robot design, we can select the correct robot diameter to complete the task and fabricate the robot at that diameter. If we must work with a specific robot diameter, we can modify the robot's launch angle and pressure so that it is more likely to succeed.

\section{Conclusion and Future Work} \label{conclusion}
In this paper, we presented a model to predict the length at which a vine robot collapses under its own weight. We validated this model through three experiments, which determined the length at which the robot would collapse at various launch angles and body diameters, as well as the pressure needed to collapse and uncollapse the robot when it was at a fixed length. We found that collapse length increases non-linearly with launch angle magnitude and linearly with diameter. We also found the pressure to collapse a robot to be lower than modeled, while the pressure to uncollapse it was higher than modeled. This supports the existence of a region where the robot's behavior is determined by the state it is in when it enters the region. This model provides a means of selecting a launch angle, body diameter, and pressure that will allow a vine robot to successfully complete a task. We demonstrated this by choosing various parameters with which a vine robot could successfully cross a gap. 

Future work will consider the material stiffness and flap effects to quantify some of the unexpected behavior observed here. We will then incorporate deflection into our model to accurately depict the path of the robot as it grows. We will also consider ways to intentionally collapse the robot, as this may be desirable in certain situations. Finally, we will add external loading to our model, generalize it for arbitrary shapes, and add in dynamic growth effects so that we can plan a non-collapsing path and robot design for any vine robot navigation task. 

\printbibliography
\end{document}